\RequirePackage{fix-cm}
\documentclass[twocolumn]{svjour3}

\smartqed

\usepackage[misc]{ifsym}
\usepackage[colorlinks,linkcolor=red,anchorcolor=green,citecolor=blue]{hyperref}
\usepackage{epstopdf}
\usepackage{graphicx}
\usepackage{float}
\usepackage{color}
\usepackage{booktabs}
\usepackage{threeparttable}
\usepackage{arydshln}
\usepackage{colortbl}
\usepackage[section]{placeins}
\usepackage{caption}
\usepackage{tabularx}
\usepackage{booktabs}
\usepackage{amsmath,amsfonts,amssymb,bm}
\usepackage{multirow}
\usepackage{diagbox}
\usepackage[ruled]{algorithm2e}
\usepackage{autobreak}
\usepackage{soul}
\soulregister\cite7
\soulregister\ref7
\usepackage[numbers]{natbib}
\usepackage{lineno,hyperref}
\usepackage{dsfont}
\usepackage{makecell,rotating}

\begin{document}

\title{RBUE: A ReLU-Based Uncertainty Estimation Method of Deep Neural Networks}

\author{
		Yufeng Xia\textsuperscript{1} \and
        Jun Zhang\textsuperscript{2} \and
        Zhiqiang Gong\textsuperscript{2} \and
        Tingsong Jiang\textsuperscript{2} \and
        Wen Yao\textsuperscript{2}
}
   
\institute{
	Wen Yao \space \Letter \at
	\email{wendy0782@126.com} \\
	\and
	Yufeng Xia \at
	\email{xiayufeng15@outlook.com} \\
	\and
	Jun Zhang \at
	\email{mcgrady150318@163.com} \\
	\and
	Zhiqiang Gong \at
	\email{gongzhiqiang13@nudt.edu.cn} \\
	\and
	Tingsong Jiang \at
	\email{tingsong@pku.edu.cn} \\
	\at 
	\textsuperscript{1} College of Aerospace Science and Engineering, National University of Defense Technology, Changsha 410073, China \\
	\at 
	\textsuperscript{2} National Innovation Institute of Defense Technology, Chinese Academy of Military Science, Beijing 100000, China \\
}

%

\date{Received: date / Accepted: date}

\maketitle

\begin{abstract}

%

Deep neural networks (DNNs) have successfully learned useful data representations in various tasks. However, assessing the reliability of these representations remains a challenge. Deep Ensemble is widely considered the state-of-the-art method which can estimate the uncertainty with higher quality, but it is very expensive to train and test. MC-Dropout is another popular method, which is less expensive but lacks the diversity of predictions. To estimate the uncertainty with higher quality in less time, we introduce a ReLU-Based Uncertainty Estimation (RBUE) method. Instead of randomly dropping some neurons of the network as in MC-Dropout or using the randomness of the initial weights of networks as in Deep Ensemble, RBUE adds randomness to the activation function module, making the outputs diverse. Under the method, we propose two strategies, MC-DropReLU and MC-RReLU, to estimate uncertainty. We analyze and compare the output diversity of MC-Dropout and our method from the variance perspective and obtain the relationship between the hyperparameters and predictive diversity in the two methods. Moreover, our method is simple to implement and does not need to modify the existing model. We experimentally validate the RBUE on three widely used datasets, CIFAR10, CIFAR100, and TinyImageNet. The experiments demonstrate that our method has competitive performance but is \textbf{more favorable in training time and memory requirements}.

\keywords{ReLU-Based \and Uncertainty estimation \and Diverse predictions \and Deep neural networks}

\end{abstract}

\section{Introduction}
The ability of deep neural networks (DNNs) to produce useful predictions is now well understood but estimating the uncertainty of these predictions remains a challenge. Deep Ensemble \cite{lakshminarayanan2017simple} and Monte-Carlo (MC) Dropout \cite{gal2016dropout} are two of the most popular methods for uncertainty estimation. Both methods can be understood by the concept of \textit{ensembles}, which use multiple models to get diverse predictions. Deep Ensemble can be seen as an explicit ensemble on multiple models, where each model is randomly initialized and trained independently using stochastic gradient descent \cite{robbins1951stochastic}. On the other hand, MC-Dropout can be seen as an implicit ensemble on a single stochastic network, where randomness is achieved by dropping different parts of weights for each input. During inference, one can run the single network multiple times with a different weight configuration to obtain a set of predictions and an uncertainty estimate. They produce diverse predictions for a given input, which is achieved by introducing stochasticity into the training or testing process and then using an aggregated measure such as variance or entropy as an uncertainty estimator.

However, both of these methods have their weakness. MC-Dropout performs significantly worse than Deep Ensemble on some uncertainty estimation tasks \cite{lakshminarayanan2017simple,ovadia2019can,gustafsson2020evaluating}. We argue that the main reason for MC-Dropout's poor performance is the high correlation between the ensemble elements that make the overall predictions insufficiently diverse. Moreover, dropping the weights randomly will result in similar weight configurations in different models obtained by sampling, consequently, less diverse predictions \cite{fort2019deep}. Deep Ensemble does not have the above problem because ensemble elements are trained independently, leading to no similar weight configurations. Despite its success, Deep Ensemble is limited in practice due to its expensive computational and memory costs, increasing linearly with the ensemble size in training and testing phases. In terms of computation, each ensemble member requires a separate neural network to forward pass its inputs. From the memory perspective, each ensemble member requires a separate copy of neural network weights, each of which can contain up to millions (sometimes billions) of parameters \cite{wen2020batchensemble}.

In this work, we introduce a ReLU-Based uncertainty estimation (RBUE) method that tackles these challenges. It builds on the intuition that the poor performance of dropout-based methods is due to the high correlation between the multiple outputs, which makes the overall predictions insufficiently diverse. Our method is designed to achieve a trade-off between reliable uncertainty estimation and an acceptable computational cost.

Inspired by \cite{xu2015empirical}, we add randomness to the activation function to get better diverse predictions than that of MC-Dropout, and the training cost is much smaller than Deep Ensemble. Under RBUE, we propose two strategies MC-DropReLU and MC-RReLU. The main difference between them is the sampling distribution of the slope of the negative semi-axis of ReLU. During training, we use a random activation function to activate the input value for each input, and the operation is as simple as the standard dropout. During testing, we run the model multiple times for each input to obtain a set of predictions and an uncertainty estimate. Our method has only one key hyperparameter: the retention rate of the activation function $q$. We evaluate our method on several natural and synthetic datasets and demonstrate that it outperforms MC-Dropout and Bayesian neural network in accuracy and uncertainty estimation quality. Compared to Deep Ensemble, our method has competitive performance but is more favorable in training time and memory requirements. Furthermore, we analyze and compare the output diversity of MC-Dropout and our method from the variance perspective and obtain the relationship between the hyperparameters in both methods and the output diversity. To summarize, the main contributions of this work are as follows:

\begin{itemize}
	\item We propose a ReLU-Based uncertainty estimation method by adding randomness to ReLU. It can generate diverse predictions to estimate model uncertainty.
	
	\item We propose two strategies MC-DropReLU and MC-RReLU, to concretely implement our method. Moreover, through a simple analysis of output variance, we provide a basis for setting hyperparameters in our method.
	
	\item We provide a comprehensive evaluation on several public datasets to verify the effect of our methods. MC-DropReLU performs better than MC-Dropout at a similar computational cost, matching Deep Ensemble at a fraction of the cost. MC-RReLU provides an idea for the concrete realization of this method.
\end{itemize}

\section{Related Work}

In what follows, we provide a brief background in model uncertainty estimation, review the best-known methods. Among them, the most prominent and practical uncertainty estimation methods are Deep Ensemble \cite{lakshminarayanan2017simple}, and MC-Dropout \cite{gal2016dropout}.

\subsection{Background}

\textit{Uncertainty Estimation} (UE) is a pivotal component to equip DNNs with the ability to know what they do not know. It generates confidence in model predictions. Epistemic (\textit{aka} model) uncertainty \cite{Gal2016Uncertainty}, as an important uncertainty, refers to uncertainty caused by a lack of knowledge. In other words, it refers to the ignorance of the DNNs, and hence to the epistemic state of the DNNs instead of any underlying random phenomenon. This uncertainty can be explained away given enough data. And it can be obtained by multiple predictions through sampling or ensemble.

\subsection{Ensemble}

Ensemble is one of the oldest tricks in machine learning literature \cite{hansen1990neural}. By combining the outputs of several models, an ensemble can achieve better performance than any of its members \cite{xie2013horizontal,huang2017snapshot,krizhevsky2009learning,perrone1992networks}. Deep Ensemble \cite{lakshminarayanan2017simple} trains multiple DNNs with different initializations and uses all the predictions for uncertainty estimation. More recently, Ovadia et al. \cite{ovadia2019can} and Gustafsson et al. \cite{gustafsson2020evaluating} independently benchmarked existing approaches to uncertainty modeling on various datasets and architectures and observed that Deep Ensemble tends to outperform Bayesian neural networks (BNNs) in both accuracy and uncertainty estimation quality. Fort et al. \cite{fort2019deep} investigated the loss landscape and postulated that variational methods only capture local uncertainty, whereas Deep Ensemble explores different global modes. It explains why Deep Ensemble generally performs better. Despite its success on benchmarks, Deep Ensemble is limited in practice due to its expensive computational costs. During training, it needs to train multiple independent networks. Moreover, during testing, it is desirable to keep all these networks in memory.

Some methods have been proposed to tackle this issue by taking a slightly different approach towards creating an ensemble. They only need a single training to get multiple models with different weight configurations. For example, Snapshot Ensemble \cite{huang2017snapshot} trains a single network and uses its parameters at $k$ different points of the training process to instantiate $k$ networks to form the target ensemble. Snapshot Ensemble cyclically varies the learning rate, enabling the single network to converge to $k$ local minima along its optimization path. Similarly, TreeNets \cite{lee2015m} also train a single network, but this network is designed to branch out into $k$ sub-networks after the first few layers. Thus, effectively every sub-network functions as a separate member of the target ensemble. 

Although these methods partially solve the problem of training time, their prediction performance and calibration scores are usually worse than standard Deep Ensemble. Furthermore, the time advantage of these methods is obtained through some training skills. However, these training skills will make it difficult to guarantee the diversity between models and the ensemble's performance.

\subsection{Dropout}

Another smart option to model uncertainty in DNNs is the use of dropout \cite{srivastava2014dropout} as a way to approximate Bayesian variational inference. The simplicity of the key idea of this formulation is one of the main reasons for its popularity. By enabling dropout in training and testing phases and making multiple forward passes through the network using the same input data, the first two moments of the predictive distribution (mean and variance) can be estimated using the output distributions of the different passes. The mean is then used as an estimate and the variance as a measure of its uncertainty. This technique is called Monte-Carlo dropout (MC-Dropout) \cite{gal2016dropout}. Furthermore, MC-Dropout has zero memory overhead compared to a single model. However, despite its success and simplicity, different predictions made by several forward passes with randomly dropped neurons seem to be overly correlated and strongly underestimate the variance. Moreover, when using MC-Dropout in practical applications, architectural choices like where to insert the dropout layers, how many to use, and the choice of dropout rate are often either empirically made or set a priori \cite{kendall2017bayesian,jungo2017towards,verdoja2019deep}, leading to possibly suboptimal performance.

\subsection{Other Methods}

In addition to the two types of methods mentioned above, several approaches based on Bayesian neural networks (BNNs) \cite{denker1990transforming,mackay1992practical,neal2012bayesian} try to estimate predictive uncertainty by imposing probability distributions over model parameters instead of using point estimates, including Markov Chain Monte Carlo (MCMC) \cite{neal2012bayesian}, Laplace approximation \cite{mackay1992bayesian} as well as recent work on variational Bayesian methods \cite{blundell2015weight}. Although BNNs provide a set of theoretical methods for uncertainty estimation, it is usually difficult to use approximate inference techniques to infer the true posteriors of the parameters. Although these techniques are theoretically grounded, Deep Ensemble and MC-Dropout often show significantly better performance in practice \cite{ovadia2019can,gustafsson2020evaluating}, in terms of both accuracy and quality of the predictive uncertainty.

\section{Method}


In this section, we describe the proposed method in detail. We begin with the formulation of the activation function framework for embedded randomness in Section \ref{3.1}. Then, two strategies of RBUE are introduced in Section \ref{3.2}. Next, we introduce how to estimate uncertainty using two strategies, including the training and testing phases in Section \ref{3.3}. Finally, we analyze the prediction diversity of our method in Section \ref{3.4}.

\subsection{Formulation of the ReLU framework with embedded randomness} \label{3.1}

This part describes the formulation of the ReLU framework with embedded randomness. Suppose $x_0$ is an input vector of an $L$-layer fully connected neural network. Let $x_l$ be the output of the $l$-th layer and $W_l$ be the weight matrix of the $l$-th layer. Biases are neglected for the convenience of presentation.
\begin{equation}
	\begin{array}{l}
		x_{l}=\left[x_{l}[1], x_{l}[2], \ldots, x_{l}[n]\right]^{T} \in \mathbb{R}^{n}, \quad W_{l} \in \mathbb{R}^{m \times n}
	\end{array}
\end{equation}

Let $x_{l+1}^{\prime}$ be the input of $(l+1)$-th activation function layer. For a standard fully connected or convolution network, the $m$-dimensional input vector can be written as
\begin{equation}
	\begin{aligned}
		x_{l+1}^{\prime} &=W_{l} x_{l} \\
		&=\left[\sum_{i=1}^{n} W_{l}[1][i] \cdot x_{l}[i], \ldots, \sum_{i=1}^{n} W_{l}[m][i] \cdot x_{l}[i]\right]^{T}
	\end{aligned}
\end{equation}

$f(\cdot)$ is the element-wise nonlinear activation operator that maps an input vector to an output vector by applying a nonlinearity on each input. We assume $f: \mathbb{R}^{m} \rightarrow \mathbb{R}^{m}$ and the output of $(l+1)$-th layer can be written as
\begin{equation} \label{sigma}
	\begin{aligned}
		x_{l+1} &=f_{l+1}\left(x_{l+1}^{\prime}\right)\\
		&= \left[\sigma_{l+1}^{1}\left(x_{l+1}^{\prime}[1]\right), \ldots, \sigma_{l+1}^{m}\left(x_{l+1}^{\prime}[m]\right)\right]^{T}
	\end{aligned}
\end{equation}

In Eqn \ref{sigma}, $\sigma$ could be a ReLU, a sigmoid, or a tanh function, but we only consider the $\sigma$ as a variant of ReLU function that is random in our paper. The randomness is given by
\begin{equation} \label{a}
	\sigma_{l+1}^{m}\left(x_{l+1}^{\prime}[m]\right)=\left\{\begin{array}{l}
		 \quad \ \ \ x_{l+1}^{\prime}[m] \ ,\ if  \ x_{l+1}^{\prime}[m] \ge 0 \\
		a_{l+1}^{m} x_{l+1}^{\prime}[m] \ ,\ if  \ x_{l+1}^{\prime}[m]<0
	\end{array}\right.
\end{equation}
where $a_{l+1}^{m}$ is a random parameter and  $a_{l+1}^{m} \sim P^{*}$. $P^{*}$ can be a continuous random distribution like uniform distribution or a discrete random distribution like Bernoulli distribution.

From Eqn \ref{a}, it can be seen that the random component in our method is mainly due to the slope of the line on the negative half-axis of the x-axis being a random number. Such a random framework can bring two benefits.

1) The same neuron will receive different activation outputs for each forward propagation of the neural network, allowing the neural network not to be overly dependent on certain neurons, thus improving model generalization.

2) Adding randomness to the activation function is less modifying to the model than other neural network modules, and it can be applied to the estimation of model uncertainty.

\subsection{Two strategies of RBUE} \label{3.2}

\begin{figure}
	\centering
	\includegraphics[scale=0.95]{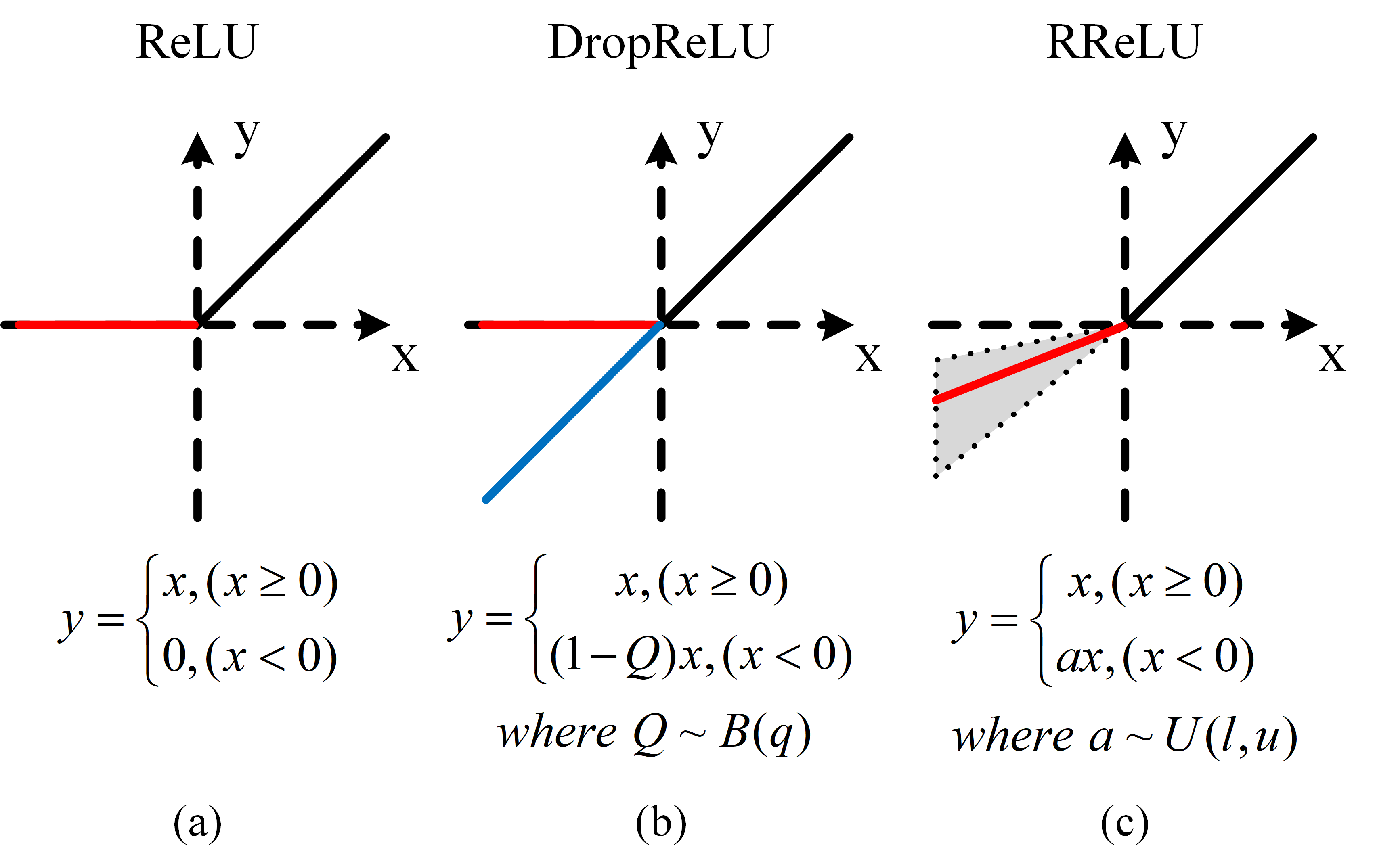}
	\caption{Three activation function. (a) is regular ReLU. (b) and (c) represent two random activation funtions, DropReLU and RReLU, respectively.}
	\label{relu}
\end{figure}

\begin{figure}
	\centering
	\includegraphics[scale=0.8]{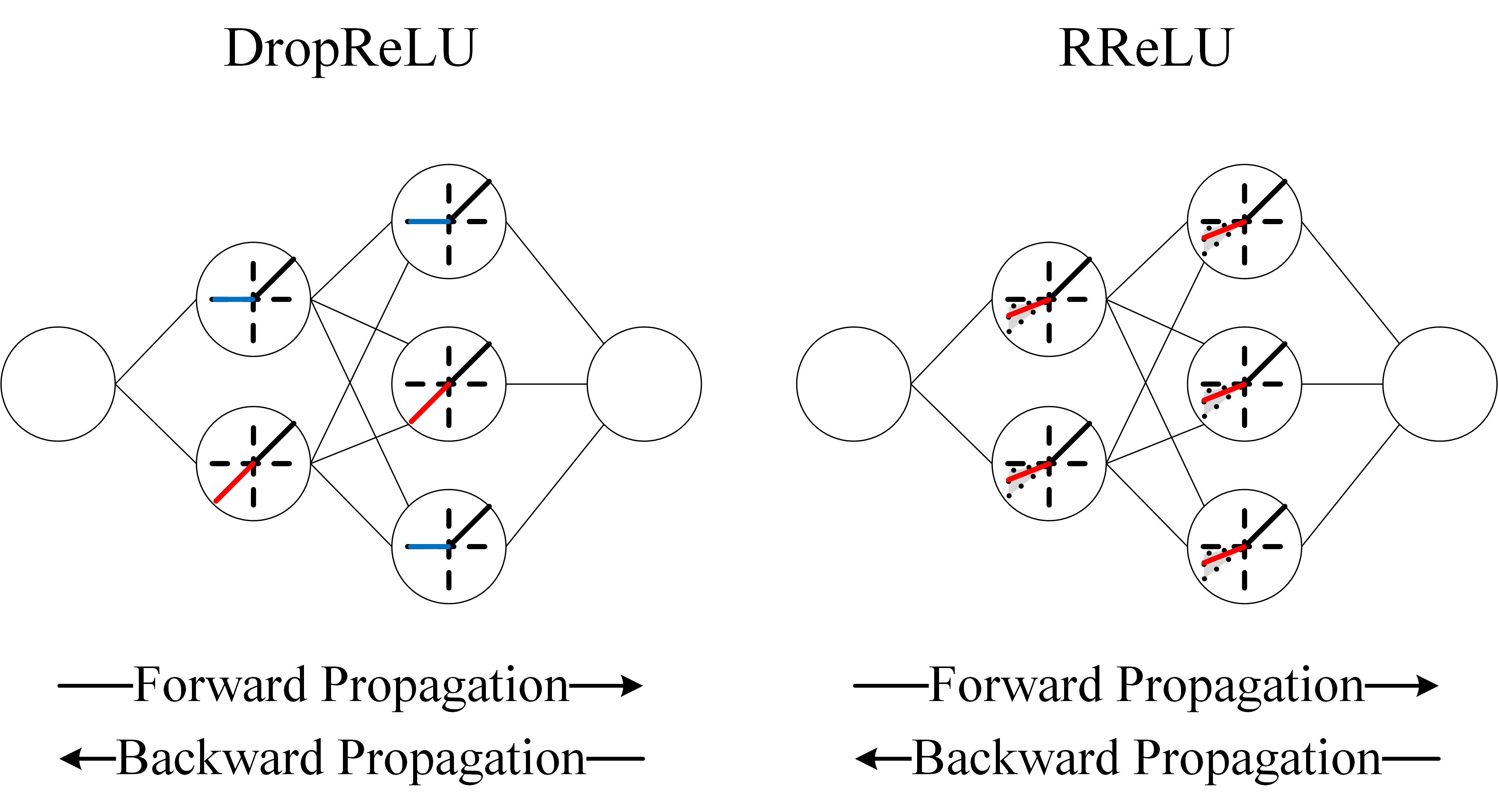}
	\caption{Neural network using DropReLU (left) or RReLU (right).}
	\label{relu3}
\end{figure}

When $P^{*}$ follows a Bernoulli distribution, we call this ReLU with embedded randomness as \textit{DropReLU}.  When $P^{*}$ follows a uniform distribution, we call this ReLU with embedded randomness as \textit{RReLU}. As shown in Figure \ref{relu}, from (a) to (c), they are ReLU, DropReLU and RReLU, respectively. Figure \ref{relu3} shows how these two strategies are used.

\begin{figure*}
	\centering
	\includegraphics[scale=1]{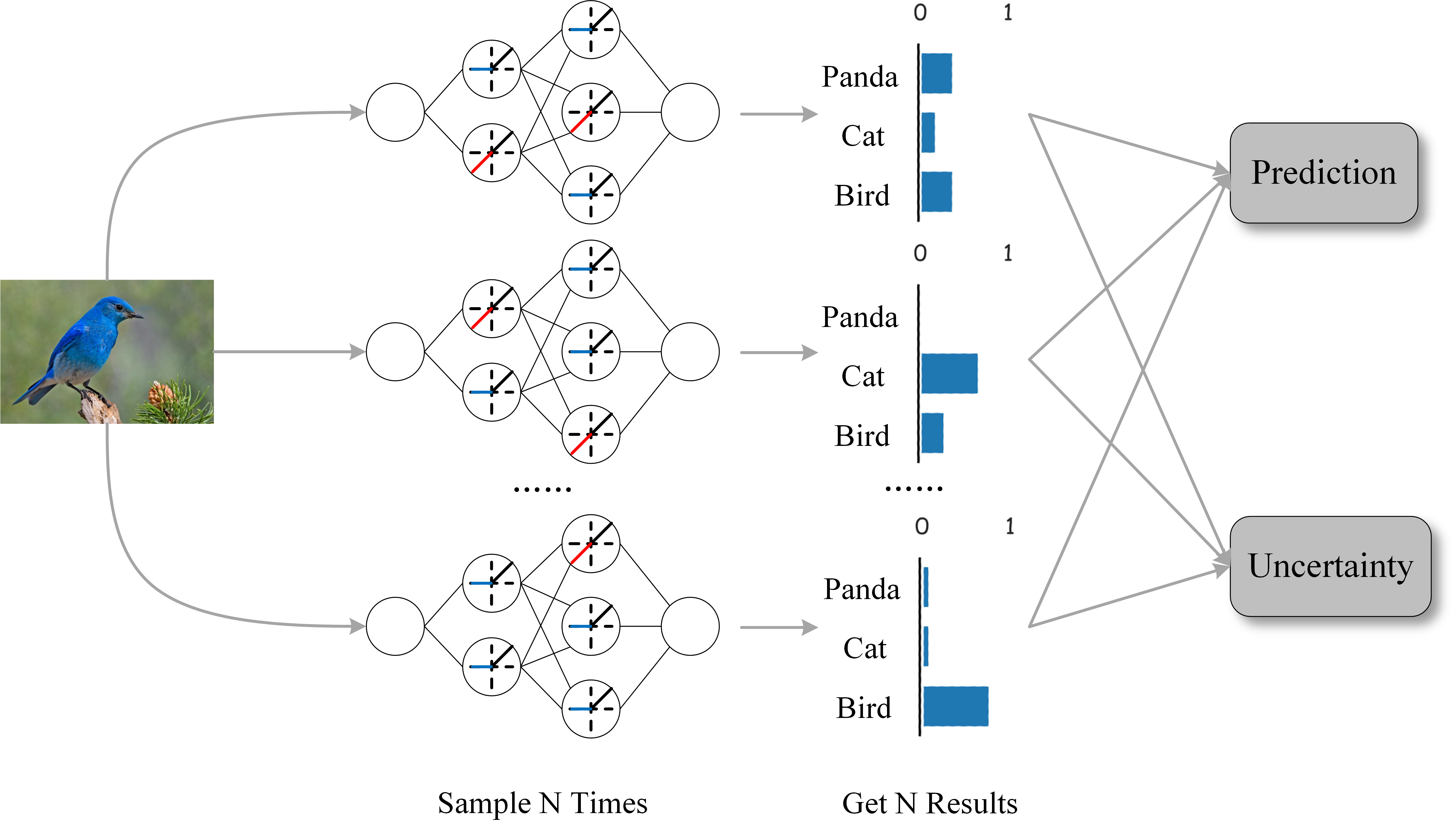}
	\caption{Framework diagram of MC-DropReLU in image classification problem. We get the final prediction and uncertainty for the same image by performing $N$ stochastic forward passes through the network. Similarly, MC-RReLU is the same framework.}
	\label{mcdroprelu}
\end{figure*}

\subsubsection{Strategy \uppercase\expandafter{\romannumeral1}: Drop Rectified Linear Unit for Uncertainty Estimation}

In this strategy, we drop the pointwise nonlinearities in $f$ randomly. Specifically, the $m$ nonlinearities $\sigma$ in the operator $f$ are kept with probability $q$ (or dropping them with probability $1-q$). The Eqn \ref{a} can be rewritten as
\begin{equation}
	\sigma_{l+1}^{m}\left(x_{l+1}^{\prime}[m]\right)=\left\{\begin{array}{l}
		\qquad \qquad \ x_{l+1}^{\prime}[m] \ ,\ if  \ x_{l+1}^{\prime}[m] \ge 0 \\
		(1-Q_{l+1}^{m}) x_{l+1}^{\prime}[m] \ ,\ if  \ x_{l+1}^{\prime}[m]<0
	\end{array}\right.
\end{equation}
where $Q_{l+1}^{m}$ is a random variable following a Bernoulli distribution $B(q)$ that takes value 1 with probability $q$ and 0 with probability $1-q$. Intuitively, when $Q=1$, then $x_{l+1} =f_{l+1}\left(x_{l+1}^{\prime}\right)=ReLU(x_{l+1}^{\prime})$, meaning all the nonlinearities in this layer are kept. When $Q=0$, then $x_{l+1} =f_{l+1}\left(x_{l+1}^{\prime}\right)=x_{l+1}^{\prime}$, meaning all the nonlinearities are dropped. The general case lies somewhere between these two limits where the nonlinearities are kept or dropped partially. At each iteration, a different realization of $Q$ is sampled from the Bernoulli distribution again. We use a combination of the above randomness and Monte Carlo method to estimate the model uncertainty.

In the experiments of this paper, we take $q$ as 0.8, 0.85, 0.9, and 0.95. Among them, $q=0.8$ and $p=0.2$ are used as a comparison to prove the analysis of variance in \ref{3.4}.

\subsubsection{Strategy \uppercase\expandafter{\romannumeral2}: Random Rectified Linear Unit for Uncertainty Estimation}

In this strategy, Random Rectified Linear Unit (RReLU) is the random version of leaky ReLU \cite{maas2013rectifier} which is first proposed and used in Kaggle National Data Science Bowl (NDSB) Competition. Although RReLU has been proposed, previous researchers only paid attention to its randomness in training to reduce the risk of overfitting. They did not pay attention to its randomness in testing that can be used to estimate model uncertainty. Moreover, this feature satisfies the framework we proposed. The highlight of RReLU is that the slope of the line on the negative half-axis of the x-axis is a random variable sampled from a uniform distribution $U(l,u)$. The Eqn \ref{a} can be rewritten as
\begin{equation}
	\sigma_{l+1}^{m}\left(x_{l+1}^{\prime}[m]\right)=\left\{\begin{array}{l}
		\quad \ \ \ x_{l+1}^{\prime}[m] \ ,\ if  \ x_{l+1}^{\prime}[m] \ge 0 \\
		a_{l+1}^{m} x_{l+1}^{\prime}[m] \ ,\ if  \ x_{l+1}^{\prime}[m]<0
	\end{array}\right.
\end{equation}
where $a_{l+1}^{m}$ is a random variable following a uniform distribution $U(l,u)$ with $l<u$ and $l,u \in [0,1)$. Suggested by the NDSB competition winner, $a_{l+1}^{m}$ is sampled from $U(\frac{1}{8},\frac{1}{3})$. We use the same configuration in this paper.

\subsection{Sampling at test time to estimate model uncertainty} \label{3.3}

As shown in Figure \ref{mcdroprelu}, we use DropReLU as an example to illustrate how to estimate model uncertainty. 

\textbf{Training phase.} The network is trained just like a regular ReLU network. The only change is to replace ReLU with one of the two random ReLUs mentioned in the previous part. Moreover, such a substitution will not affect the generalization of the model, nor will it affect the training time of the model.

\textbf{Testing phase.} The connection of the network is the same as in the training phase and does not require any changes. Keeping the above two ReLUs enabled during test time allows us to perform multiple forward passes to get multiple networks with different parameters. We refer to this Monte Carlo estimation as \textit{MC-DropReLU} (\textit{MC-RReLU}). In practice, this is equivalent to performing $N$ stochastic forward passes through the network and averaging the results. As we can see from Figure \ref{mcdroprelu}, the final prediction result and predictive uncertainty are derived from the mean and entropy of $N$ sets of outputs, just like the operation in MC-Dropout \cite{gal2016dropout}.

\subsection{Analysis of variance of predictions} \label{3.4}

In this part, we prove that our method is better than MC-Dropout in prediction diversity by variance analysis. To simplify the analysis, we only analyze one layer in the neural network and ignore the bias. To this end, suppose that layer $i$ is a fully connected layer, $x$ is the output of layer $i$ and the input of the Dropout layer or DropReLU layer after layer $i$.

For the Dropout layer, its output can be formulated as 

\begin{equation}
	f_{Dropout}=\sum_{k=1}^{K} P_{k} \cdot x_{k}
\end{equation}
where $P_{k} \sim B(p)$ and it takes value 0 with probability $p$ and 1 with probability $1-p$. $K$ represents the number of neurons in layer $i$. The variance of the output of Dropout layer is

\begin{equation}
	\operatorname{Var}(f_{Dropout})=\operatorname{Var}\left(\sum_{k=1}^{K} P_{k} \cdot x_{k} \right) =p(1-p) \sum_{k=1}^{K} x_{k}^{2}
\end{equation}

For the DropReLU layer, its output can be formulated as

\begin{equation}
	f_{DropReLU}=\sum_{k=1}^{K} [(1-Q_{k}) \cdot x_{k} + Q_{k} \cdot ReLU(x)]
\end{equation}
where $Q_{k} \sim B(q)$ and it takes value 0 with probability $1-q$ and 1 with probability $q$. $K$ represents the number of neurons in layer $i$. The variance of the output of DropReLU layer is

\begin{equation}
	\begin{aligned}
		&\operatorname{Var}(f_{DropReLU})\\
		=&\operatorname{Var}\left(\sum_{k=1}^{K} [(1-Q_{k}) \cdot x_{k} + Q_{k} \cdot ReLU(x)] \right) \\
		=&q(1-q) \sum_{k=1}^{K} x_{k}^{2} + \epsilon 
	\end{aligned}
\end{equation}
where $\epsilon=\operatorname{Var}\left(\sum_{k=1}^{K} Q_{k} \cdot \operatorname{ReL} U\left(x_{k}\right)\right)>0$. This cannot be calculated, but it can be guaranteed that it is always greater than 0.

Through theoretical analysis, it can be known that when $q \le 1-p$, the variance of the output of DropReLU is always greater than the variance of the output of Dropout, which also means that the diversity of the output of DropReLU is better than that of Dropout. This conclusion guides us in setting up the experiment's hyperparameters, and the experimental results also prove this conclusion. When $q > 1-p$, because $\epsilon$ cannot be calculated, we still need to look at the experimental results.

\section{Experiments}

In this section, we show the superiority of our proposed method by several experiments. We use these experiments to answer the following questions:
\begin{itemize}
	\item[Q1.] How accurate are the predictions, and how reliable is the uncertainty estimated by MC-DropReLU and MC-RReLU under \textbf{clean} datasets compared to other baselines?
	\item[Q2.] How accurate are the predictions, and how reliable is the uncertainty estimated by MC-DropReLU and MC-RReLU under \textbf{corruptional} datasets (a kind of out-of-distribution datasets) compared to other baselines?
	\item[Q3.] How diverse of neural networks in MC-DropReLU and MC-RReLU compared with baselines?
	\item[Q4.] What effect does the position and configuration of random ReLU appearing in the neural network on the predictive accuracy and uncertainty?
\end{itemize}

\begin{figure*}
	\centering
	\includegraphics[width=\textwidth]{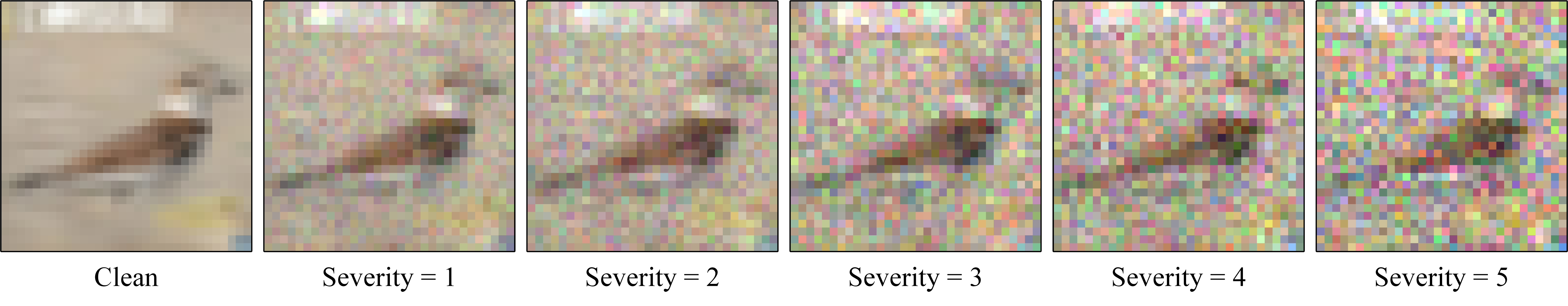}
	\caption{Examples of CIFAR-10 images corrupted by shot noise, at severities of 0 (uncorrupted image) through 5 (maximum corruption included in CIFAR-10-C).}
	\label{cifar10c}
\end{figure*}

\subsection{Preparation} \label{4.1}

\subsubsection{Dataset}

\textbf{CIFAR10} and \textbf{CIFAR100} consists of 60000 32$\times$32 colour images in 10 and 100 classes, with 6000 and 600 images per class, respectively. There are 50000 training images and 10000 test images. We adopt a standard data augmentation scheme that is widely for these two datasets \cite{he2016deep,huang2016deep,larsson2016fractalnet,lin2013network,romero2014fitnets,lee2015deeply,springenberg2014striving,srivastava2015training}. For preprocessing, we normalize the data using the channel means and standard deviations.

\textbf{TinyImageNet} dataset consists of 120000 64$\times$64 color images in 200 classes, with 600 images per class. There are 100000 training images, 10000 test images and 10000 validation images.

\textbf{CIFAR10-C} and \textbf{TinyImageNet-C} datasets consist of 19 diverse corruption types applied to validation images of CIFAR10 and TinyImageNet. The corruptions are drawn from four main categories—noise, blur, weather, and digital. Each corruption type has five levels of severity since corruption can manifest itself at varying intensities. Figure \ref{cifar10c} gives an example of the five different severity levels for shot noise. In our experiments, we test networks with CIFAR10-C and TinyImageNet-C images, but networks should not be trained on CIFAR10-C and TinyImageNet-C. Networks should be trained on datasets such as CIFAR10 and TinyImageNet. Overall, the CIFAR10-C and TinyImageNet-C datasets consist of 95 corruptions, and all are applied to CIFAR10 and TinyImageNet validation images for testing a pre-existing network.

\subsubsection{Experiment setting}

In this part, we will explain our experimental setup in detail.

The VGG \cite{simonyan2014very}, ResNet \cite{he2016deep} and DenseNet \cite{huang2017densely} models are implemented using Pytorch 1.7. All the networks are trained using stochastic gradient descent (SGD) \cite{robbins1951stochastic}. On CIFAR10 and CIFAR100, we train using batch size 128 for 200 epochs. The initial learning rate is set to 0.1 and divided by ten at 45\%, 67.5\%, and 90\% of the training epochs. On TinyImageNet, we train using batch size 100 for 150 epochs. The initial learning rate is set to 0.01 and divided by ten at 60\% and 90\% of the training epochs. We use a weight decay of $10^{-4}$ and a Nesterov momentum \cite{sutskever2013importance} of 0.9. For the stochastic method, we average 100 sample predictions to yield a predictive distribution.

All experiments are run on the same server with NVIDIA RTX 3090 GPU.

\begin{table*}[htbp]\scriptsize
	\centering
	\caption{Comparison over \textbf{CIFAR10} with VGG-13, ResNet-18 and Densenet-121 models on three metrics. In terms of model size, Deep Ensemble is four times that of all other methods, which means its storage space is four times that of all other methods. \textbf{Bold} numbers indicate the best way to balance the three metrics.}
	\begin{tabular}{l|ccc|ccc|ccc}
		\toprule
		Model & \multicolumn{3}{c|}{VGG-13} & \multicolumn{3}{c|}{ResNet-18} & \multicolumn{3}{c}{DenseNet-121} \\
		\midrule
		\midrule
		Approach & Accuracy & ECE   & Training Time & Accuracy & ECE   & Training Time & Accuracy & ECE   & Training Time \\
		&  (\%, $\uparrow$) &  ($\downarrow$)  & ($\downarrow$) & (\%, $\uparrow$) & ($\downarrow$)  & ($\downarrow$)& (\%, $\uparrow$) & ($\downarrow$)  & ($\downarrow$) \\
		\midrule
		\midrule
		Single & 93.85 & 0.05  & 1.4h  & 95.01 & 0.04  & 1.7h  & 94.20  & 0.05  & 2.9h \\
		MC-Dropout(0.2) & 93.78 & 0.04  & 1.4h  & 94.99 & 0.03  & 1.9h  & 94.05 & 0.04  & 3.4h \\
		MC-Dropout(0.5) & 93.74 & 0.04  & 1.4h  & 95.26 & 0.03  & 1.9h  & 94.16 & 0.04  & 3.4h \\
		Deep Ensemble(4) & 94.98 & 0.03  & 5.6h  & 96.12 & 0.02  & 6.8h  & 95.27 & 0.03  & 11.6h \\
		SVI   & 92.80  & 0.05  & 1.5h  & 94.42 & 0.04  & 1.9h  & 93.48 & 0.05  & 3.6h \\
		\midrule
		\midrule
		MC-RReLU & 93.58 & 0.03  & 1.4h  & 94.68 & 0.03  & 1.9h  & 93.73 & 0.03  & 3.5h \\
		MC-DropReLU(0.8) & 93.37 & 0.04  & 1.5h  & 94.33 & 0.02  & 2.0h  & 93.63 & 0.03  & 4.3h \\
		MC-DropReLU(0.85) & 93.46 & 0.04  & 1.5h  & 94.64 & 0.03  & 2.0h  & 93.84 & 0.03  & 4.3h \\
		MC-DropReLU(0.9) & \textbf{93.89} & \textbf{0.03}  & \textbf{1.5h}  & 94.86 & 0.03  & 2.0h  & 93.89 & 0.03  & 4.3h \\
		MC-DropReLU(0.95) & 93.76 & 0.04  & 1.5h  & \textbf{95.32} & \textbf{0.02}  & \textbf{2.0h}  & \textbf{94.16} & \textbf{0.03}  & \textbf{4.3h} \\
		\bottomrule
	\end{tabular}%
	\label{C10clean}%
\end{table*}%

\begin{table*}[htbp]\scriptsize
	\centering
	\caption{Comparison over \textbf{CIFAR100} with VGG-13, ResNet-18 and Densenet-121 models on three metrics. In terms of model size, Deep Ensemble is four times that of all other methods, which means its storage space is four times that of all other methods. \textbf{Bold} numbers indicate the best way to balance the three metrics.}
	\begin{tabular}{l|ccc|ccc|ccc}
		\toprule
		Model & \multicolumn{3}{c|}{VGG-13} & \multicolumn{3}{c|}{ResNet-18} & \multicolumn{3}{c}{DenseNet-121} \\
		\midrule
		\midrule
		Approach & Accuracy & ECE   & Training Time & Accuracy & ECE   & Training Time & Accuracy & ECE   & Training Time \\
		&  (\%, $\uparrow$) &  ($\downarrow$)  & ($\downarrow$) & (\%, $\uparrow$) & ($\downarrow$)  & ($\downarrow$)& (\%, $\uparrow$) & ($\downarrow$)  & ($\downarrow$) \\
		\midrule
		\midrule
		Single & 74.13 & 0.13  & 1.4h  & 77.01 & 0.11  & 1.7h  & 75.74 & 0.12  & 2.9h \\
		MC-Dropout(0.2) & 74.13 & 0.11  & 1.4h  & 77.19 & 0.08  & 1.9h  & 75.57 & 0.09  & 3.4h \\
		MC-Dropout(0.5) & 74.20 & 0.11  & 1.4h  & 77.25 & 0.09  & 1.9h  & 75.10 & 0.10   & 3.4h \\
		Deep Ensemble(4) & 75.61 & 0.08  & 5.6h  & 79.45 & 0.06  & 6.8h  & 77.48 & 0.06  & 11.6h \\
		SVI   & 72.12 & 0.11  & 1.8h  & 74.11 & 0.11  & 1.9h  & 72.11 & 0.14  & 3.1h \\
		\midrule
		\midrule
		MC-RReLU & \textbf{74.31}  & \textbf{0.09}  & \textbf{1.4h}  & 77.05 & 0.09  & 1.9h  & 75.63 & 0.07  & 3.4h \\
		MC-DropReLU(0.8) & 73.90  & 0.11  & 1.5h  & 76.36 & 0.09  & 2.0h  & 74.86 & 0.06  & 3.9h \\
		MC-DropReLU(0.85) & 73.99 & 0.11  & 1.5h  & 77.00    & 0.09  & 2.0h  & \textbf{75.48} & \textbf{0.06}  & \textbf{3.9h} \\
		MC-DropReLU(0.9) & 73.94 & 0.11  & 1.5h  & 77.76 & 0.09  & 2.0h  & 76.50  & 0.08  & 3.9h \\
		MC-DropReLU(0.95) & 74.41 & 0.11  & 1.5h  & \textbf{78.10}  & \textbf{0.07}  & \textbf{2.0h}  & 76.27 & 0.09  & 3.9h \\
		\bottomrule
	\end{tabular}%
	\label{C100clean}%
\end{table*}%

\subsubsection{Metrics}


We measure classification accuracy, calibration score (ECE \cite{friedman2001elements, guo2017calibration, naeini2015obtaining}), model size, training time, and model diversity. (The arrow behind the metric represents which direction is better.)

\textbf{Expected Calibration Error (ECE $\downarrow$)}. Let $B_{m}$ be a set of indices of test examples whose prediction scores for the ground-truth labels fall into interval $\left(\frac{m-1}{M}, \frac{m}{M}\right]$ for $m \in\{1, \ldots M\}$, where $M$ (= 30) is the number of bins. ECE is formally defined by
\begin{equation}
	\text {ECE}=\sum_{m=1}^{M} \frac{\left|B_{m}\right|}{n}\left|\operatorname{acc}\left(B_{m}\right)-\operatorname{conf}\left(B_{m}\right)\right|
\end{equation}
where $n$ is the number of the test samples. Also, accuracy and 
confidence of each bin are given by
\begin{equation}
	\begin{aligned}
		\operatorname{acc}\left(B_{m}\right) &=\frac{1}{\left|B_{m}\right|} \sum_{i \in B_{m}} \mathds{1}\left(\hat{y}_{i}=y_{i}\right) \\
		\operatorname{conf}\left(B_{m}\right) &=\frac{1}{\left|B_{m}\right|} \sum_{i \in B_{m}} p_{i}
	\end{aligned}
\end{equation}
where $\mathds{1}$ is an indicator function, $\hat{y}_{i}$ and $y_{i}$ are predicted and true label of the $i^{th}$ example and $p_{i}$ is its predicted confidence. We note that a low value for this calibration score means that the network is well-calibrated.

\textbf{Model Size and Training Time $\downarrow$}. A major motivation for our method is to match the performance of Deep Ensembles while using a smaller model that requires significantly less memory. Therefore, we use the total number of weights that parameterize our models as a proxy for that. In addition to the model size, we also report the total training time used to train any particular model.

\textbf{Model Diversity $\uparrow$}. The diversity between models plays an important role in the estimation method of model uncertainty. In this paper, we use two methods to measure model diversity: Jensen-Shannon Divergence (JSD) \cite{MLSYS2020_3ef81541} and Disagreement of Predictions (DIS) \cite{fort2019deep}. They both reflect the diversity of models by measuring the inconsistency between different results obtained by different models for the same input.

%

\subsubsection{Baselines}

We compare our methods (\romannumeral1) \textit{MC-DropReLU}: Monte-Carlo DropReLU with different rate $q (=0.8,0.85,0.9,0.95)$ and (\romannumeral2) \textit{MC-RReLU}: Monte-Carlo RReLU with upper bound $u (=\frac{1}{3})$ and lower bound $l (=\frac{1}{8})$, to (a) \textit{Single}: maximum softmax probability of single model \cite{hendrycks2016baseline}, (b) \textit{MC-Dropout}: Monte-Carlo Dropout with different rate $p (=0.2,0.5)$ \cite{gal2016dropout}, (c) \textit{Deep Ensemble}: Ensembles of $M$ networks trained independently on the entire dataset using random initialization \cite{lakshminarayanan2017simple} (we set $M$ = 4 in experiments below), (d) \textit{SVI}: Stochastic Variational Bayesian Inference for deep learning \cite{wu2018deterministic}.

\subsection{CIFAR10/CIFAR100 and CIFAR10-C} \label{4.2}

In this part, we focus on Question 1 and Question 2. Table \ref{C10clean} and Table \ref{C100clean} present accuracy and ECE for several combinations of network architectures and CIFAR10/CIFAR100 datasets. Higher accuracy means better generalization performance, and lower ECE means higher quality predictive uncertainty.

We train the corresponding models with the corresponding methods and then evaluate multiple metrics separately. The results presented in both Table \ref{C10clean} and Table \ref{C100clean} indicate that our proposed RBUE framework can produce reliable uncertainty estimates on par with Deep Ensemble at a significantly lower computational cost. Even if our methods do not achieve the same effect as Deep Ensemble, they are the closest. It can also be seen from Table \ref{C10clean} and Table \ref{C100clean} that MC-DropReLU outperforms MC-Dropout in all metrics regardless of the value of $q$. However, there is still a slight difference in the effect of different $q$ for different models. When $q=0.95$, the ResNet and DenseNet models will have good accuracy and calibration scores, while the VGG model will have better accuracy and calibration scores when $q=0.9$. As shown in Table \ref{C10clean} and Table \ref{C100clean}, it is worth mentioning that when $p=0.2$ and $q=0.8$, $q \le 1-p$ is satisfied. The experimental results show that the uncertainty quality of MC-DropReLU is better than that of MC-Dropout, which verifies the analysis in \ref{3.4}. Although SVI has theoretical support, experiments show that the accuracy and ECE of this method deteriorate as the model and dataset become more complex, which is why SVI method is not used much in vision tasks.

The model size and training time in Table \ref{C10clean} and Table \ref{C100clean} also reflect the advantages of our method. Our method does not add additional parameters compared to a single model and MC-Dropout, so the space complexity is the same as MC-Dropout and less than Deep Ensemble. This is why our methods can replace MC-Dropout with no additional cost. On the other hand, in terms of training time, our methods are slightly slower than MC-Dropout. We argue that the main reason for this is that the sampling on the random ReLU is slower than the sampling on dropout. However, the overall time is still much faster than Deep Ensemble.

\begin{figure}
	\centering
	\includegraphics[scale=1]{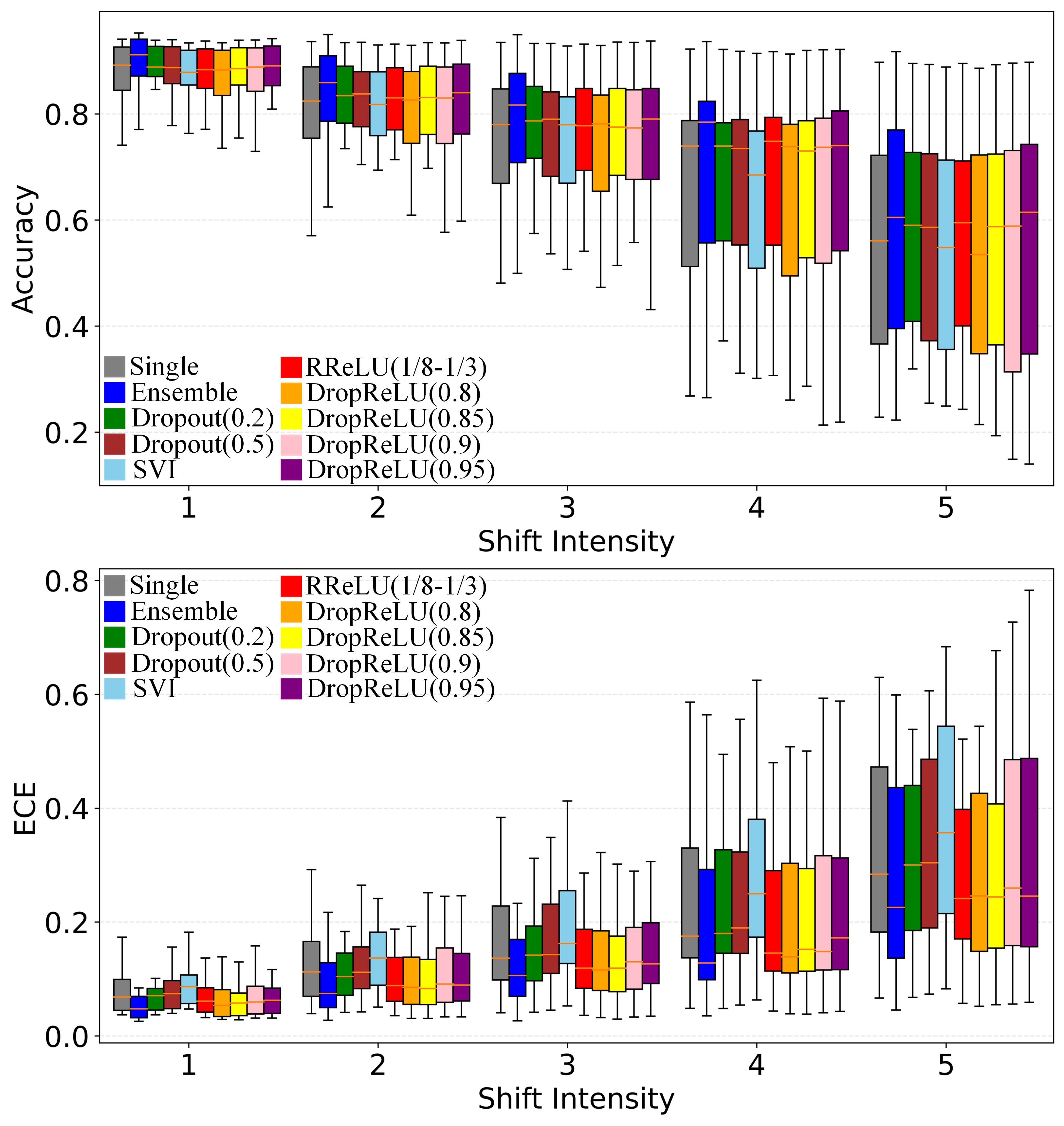}
	\caption{\textbf{CIFAR-10 results on corrupted images.} Accuracy and uncertainty metric under distributional shift: a detailed comparison of accuracy and ECE under all types of corruptions on CIFAR10 with \textbf{DenseNet-121} model.}
	\label{densenet-cifar10-c}
\end{figure}

The current neural networks are too confident about their prediction results, proposed and confirmed in \cite{guo2017calibration}. This will result in the model making a confident judgment on the data it has never seen before, but obviously, this judgment is wrong. The more confident the model is, the more it will feel that everything is certain, and therefore it will not be able to estimate high-quality uncertainty. Therefore, it is essential to evaluate the model’s calibration metrics on out-of-distribution inputs for uncertainty estimation. Following \cite{ovadia2019can}, we evaluate model accuracy and ECE on a corrupted version of CIFAR10 \cite{hendrycks2018benchmarking}. Namely, we consider 19 different ways to artificially corrupted the images and five different levels of severity for each of those corruptions.

We report our results in Figure \ref{densenet-cifar10-c}. We show the mean on the test set for each method and summarize the results on each intensity of shift with a box plot. Each box shows the quartiles summarizing the results across all 19 types of shift, while the error bars indicate the min and max across different shift types. We test six different approaches: a single network, MC-Dropout, Deep Ensemble, SVI, MC-DropReLU, and MC-RReLU. Unsurprisingly, as the severity of the perturbations increases, the advantages of our methods are becoming more obvious. Our methods perform on par with Deep Ensemble and consistently outperform MC-Dropout and SVI.

\begin{figure}
	\centering
	\includegraphics[scale=1]{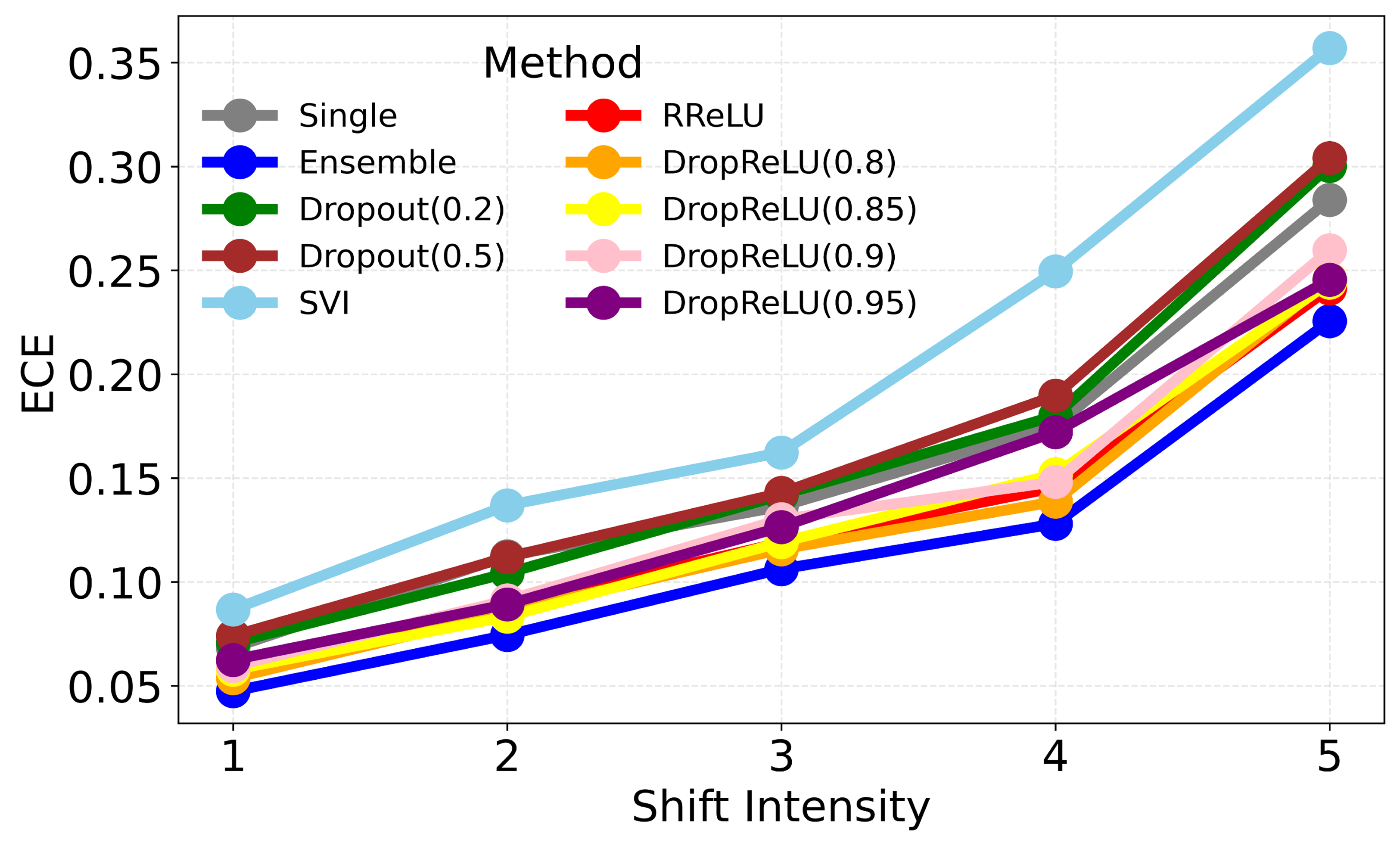}
	\caption{\textbf{CIFAR10 ECE.} ECE is a function of severity of image corruptions. Each curve represents the median of each method under different noise intensities in Figure \ref{densenet-cifar10-c}.}
	\label{densenet-cifar10-c-ece}
\end{figure}

In Figure \ref{densenet-cifar10-c-ece}, We choose the median of all the box plots in Figure \ref{densenet-cifar10-c} to compare the ECE of different methods more intuitively. Although the ECE of Deep Ensemble is the lowest under different noise intensities, our methods are the closest to Deep Ensemble among the remaining methods. Single model, MC-Dropout, and SVI all have higher ECE than our methods.

\subsection{TinyImageNet and TinyImageNet-C} \label{4.3}

\begin{table*}[htbp]\scriptsize
	\centering
	\caption{Comparison over \textbf{TinyImageNet} with VGG-13, ResNet-18 and Densenet-121 models on three metrics. In terms of model size, Deep Ensemble is four times that of all other methods, which means that its storage space is also four times that of all other methods. \textbf{Bold} numbers indicate the best way to balance the three metrics.}
	\begin{tabular}{l|ccc|ccc|ccc}
		\toprule
		Model & \multicolumn{3}{c|}{VGG-13} & \multicolumn{3}{c|}{ResNet-18} & \multicolumn{3}{c}{DenseNet-121} \\
		\midrule
		\midrule
		Approach & Accuracy & ECE   & Training Time & Accuracy & ECE   & Training Time & Accuracy & ECE   & Training Time \\
		&  (\%, $\uparrow$) &  ($\downarrow$)  & ($\downarrow$) & (\%, $\uparrow$) & ($\downarrow$)  & ($\downarrow$)& (\%, $\uparrow$) & ($\downarrow$)  & ($\downarrow$) \\
		\midrule
		\midrule
		Single & 55.93 & 0.22  & 3.5h  & 62.08 & 0.12  & 5.9h  & 62.82 & 0.06  & 8.6h \\
		MC-Dropout(0.2) & 56.57 & 0.15  & 4.0h  & 61.60  & 0.09  & 6.5h  & 63.20  & 0.05  & 10.3h \\
		MC-Dropout(0.5) & 56.90  & 0.20   & 4.0h  & 61.70  & 0.10   & 6.5h  & 62.00    & 0.05  & 10.3h \\
		Deep Ensemble(4) & 58.36 & 0.11  & 14.0h & 67.34 & 0.05  & 23.6h & 65.53 & 0.04  & 34.4h \\
		SVI   & 54.21 & 0.21  & 4.0h  & 60.05 & 0.12  & 6.0h  & 60.32 & 0.07  & 9.0h \\
		\midrule
		\midrule
		MC-RReLU & \textbf{57.38} & \textbf{0.13}  & \textbf{4.3h}  & 63.60  & 0.07  & 6.6h  & 63.01 & 0.04  & 10.4h \\
		MC-DropReLU(0.8) & 56.67 & 0.14  & 4.3h  & \textbf{64.89} & \textbf{0.04}  & \textbf{7.7h}  & 63.39 & 0.04  & 14.3h \\
		MC-DropReLU(0.85) & 56.65 & 0.15  & 4.3h  & 64.32 & 0.08  & 7.7h  & 63.43 & 0.05  & 14.3h \\
		MC-DropReLU(0.9) & 56.96 & 0.14  & 4.3h  & 64.76 & 0.05  & 7.7h  & \textbf{63.50}  & \textbf{0.04}  & \textbf{14.3h} \\
		MC-DropReLU(0.95) & 57.05 & 0.14  & 4.3h  & 64.28 & 0.08  & 7.7h  & 63.84 & 0.05  & 14.3h \\
		\bottomrule
	\end{tabular}%
	\label{imagenetclean}%
\end{table*}%

In this part, we focus on Question 1 and Question 2. Table \ref{imagenetclean} presents accuracy and ECE for several combinations of network architectures and TinyImageNet datasets. Higher accuracy means better generalization performance, and lower ECE means higher quality predictive uncertainty.

\begin{figure}
	\centering
	\includegraphics[scale=1]{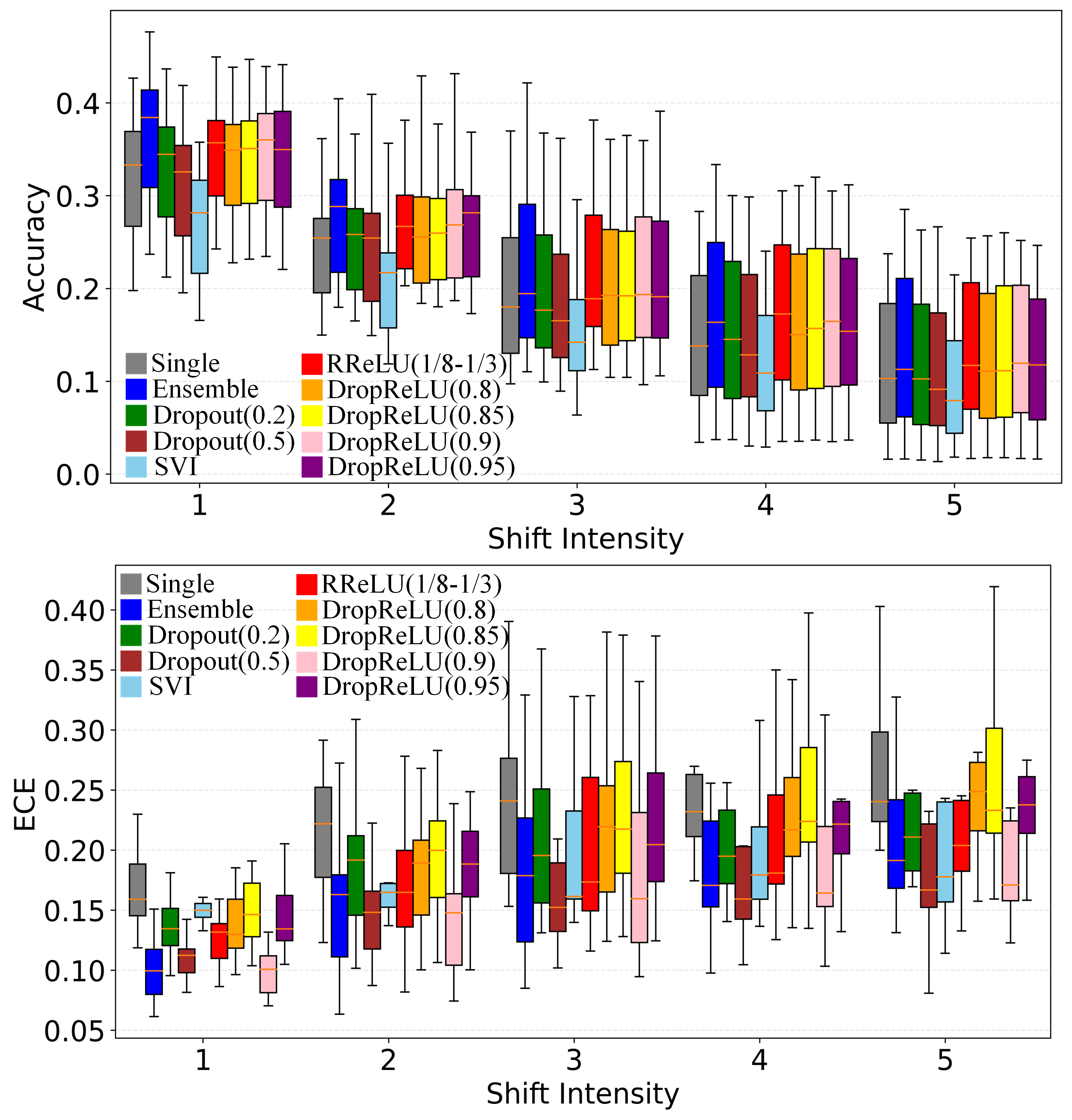}
	\caption{\textbf{TinyImageNet results on corrupted images.} Accuracy and uncertainty metric under distributional shift: a detailed comparison of accuracy and ECE under all types of corruptions on TinyImageNet with \textbf{DenseNet-121} model.}
	\label{densenet-tiny-c}
\end{figure}

\begin{figure}
	\centering
	\includegraphics[scale=1]{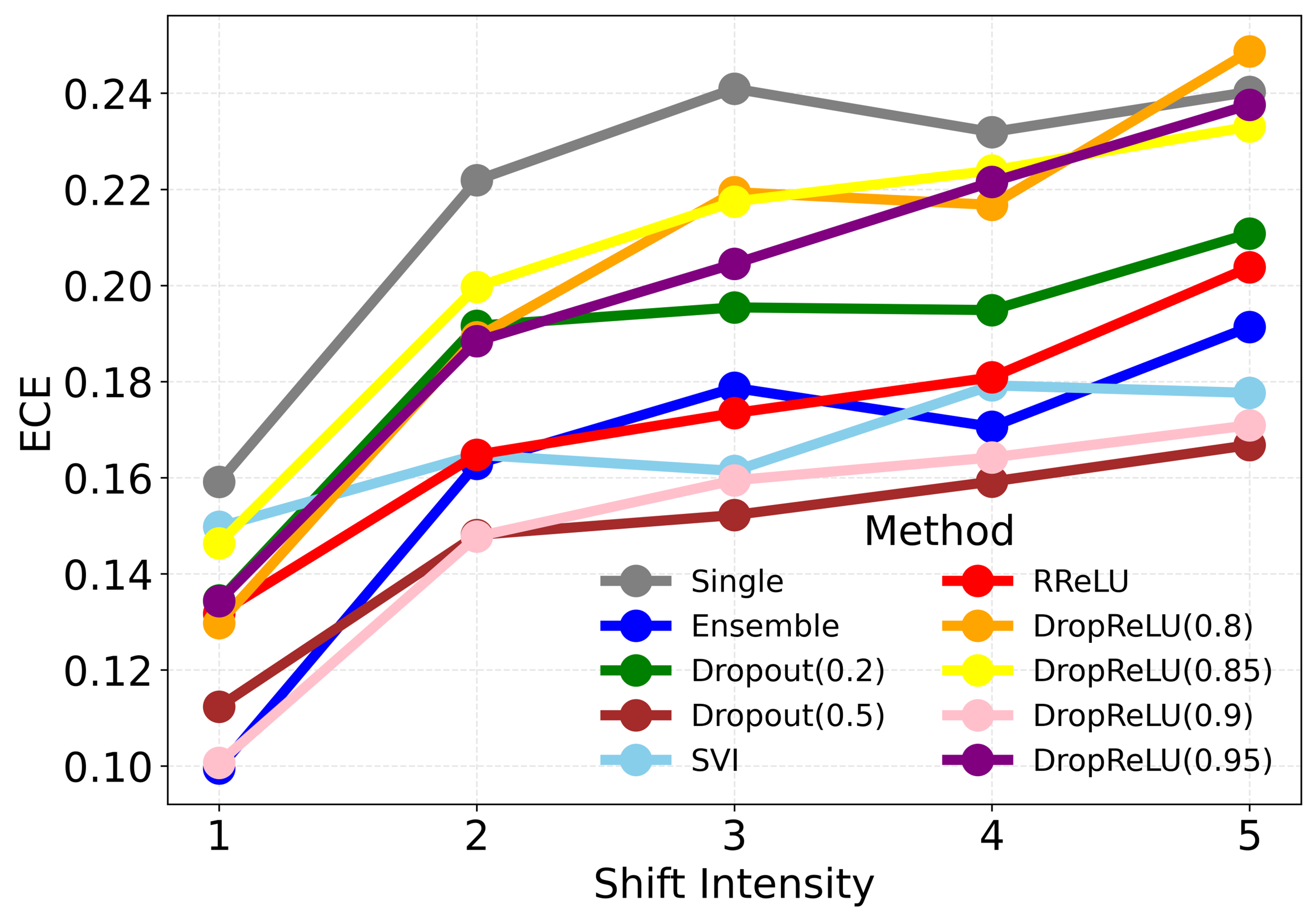}
	\caption{\textbf{TinyImageNet ECE.} ECE is a function of severity of image corruptions. Each curve represents the median of each method under different noise intensities in Figure \ref{densenet-tiny-c}.}
	\label{densenet-tiny-c-ece}
\end{figure}

We follow the same evaluation protocol as in Section \ref{4.2} and report our results on the original images in Table \ref{imagenetclean} and on the corrupted ones in Figure \ref{densenet-tiny-c}. As shown in Table \ref{imagenetclean}, the performance of our methods is similar to Deep Ensemble and significantly better than MC-Dropout in terms of accuracy and ECE on both ResNet-18 and DenseNet-121 models. However, the performance on VGG-13 model is slightly worse. We argue that the reason is that VGG model has poor generalization ability for large datasets. Note that our methods achieve these results with a training time and memory consumption four times smaller than that of Deep Ensemble and nearly the same as that of a single model. It is also worth mentioning that when $p=0.2$ and $q=0.8$, $q \le 1-p$ is satisfied. The experimental results show that the uncertainty quality of MC-DropReLU is better than that of MC-Dropout, which verifies the analysis in \ref{3.4}.

In Figure \ref{densenet-tiny-c-ece}, We choose the median of all the box plots in Figure \ref{densenet-tiny-c} to compare the ECE of different methods more intuitively. On TinyImageNet, the ECE gap between methods is more obvious than on CIFAR10. Among them, our method MC-DropReLU(0.9) exceeds Deep Ensemble in ECE at all noise intensities, showing that our methods are also applicable to large datasets.

\begin{figure*}
	\centering
	\includegraphics[width=\textwidth]{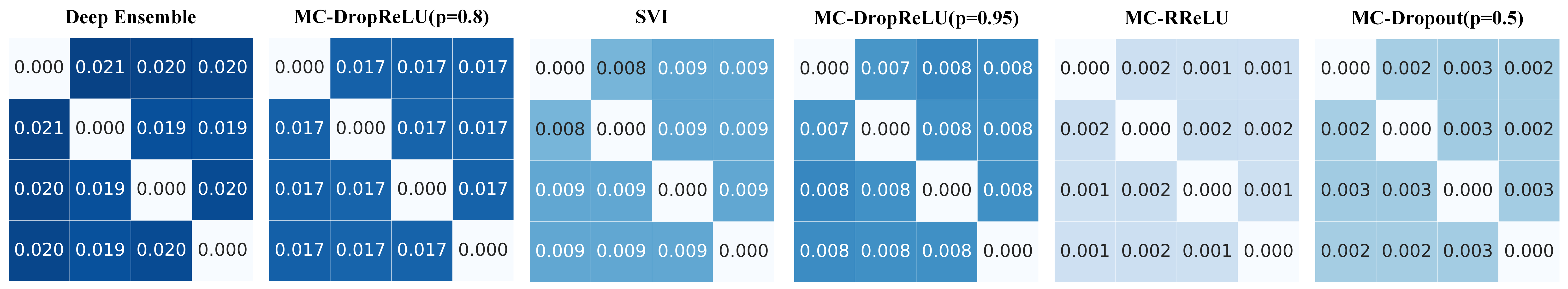}
	\caption{Using Jensen-Shannon Divergence (JSD) to characterize the diversity of ResNet-18 models under the five methods.}
	\label{js}
\end{figure*}

\subsection{Diversity analysis} \label{4.4}

In this part, we focus on Question 3. We know that diversity among models is important in uncertainty estimation. Less correlated ensembles of models deliver better performance, produce more accurate predictions \cite{hansen1990neural, perrone1992networks}, and demonstrate lower calibration error \cite{ovadia2019can}. In this paper's sampling-based uncertainty estimation method, the diversity among models represents the diversity among multiple predictions by sampling. Better diversity represents more comprehensive information captured by the multiple models obtained by sampling, which leads to a higher quality of the uncertainty estimates. In this paper, we use two evaluation methods to measure the diversity of our proposed method compared to the baseline.

\subsubsection{Divergence of predictions} \label{4.4.1}

Our goal is to see how the correlation between the different models obtained by sampling. Letting $Y_{i j}$ be the softmax output of model $i$ obtained by sampling on test input $j$, and we can think of it as a probability distribution, we then estimate Jensen-Shannon Divergence (JSD) between $Y_{i j}$ and $Y_{i^{\prime} j}$ for each $i$, $i^{\prime}$ and $j$. We then average across all test examples to get an average value for a model instead of one for each test example. Figure \ref{js} shows the results. The value corresponding to the $i$ th row and $i^{\prime}$ th column in each picture means the JSD of model $i$ and model $i^{\prime}$. Because JSD is symmetrical, the matrices in the figure are all symmetrical. JSD, Mena-JSD, and Max-JSD are formally defined by
\begin{equation}
	\text{JSD}\left(Y_{i}, Y_{i^{\prime}}\right)=\frac{1}{n} \sum_{j=1}^{n} \text{JSD}\left(Y_{i j}, Y_{i^{\prime} j}\right)
\end{equation}
\begin{equation}
	\text{Mean-JSD}=\frac{1}{6}\sum_{i=1}^{3}  \sum_{i^{\prime} =i+1}^{4} \text{JSD}\left ( Y_{i}, Y_{i^{\prime}}\right ) 
\end{equation}
\begin{equation}
	\text{Max-JSD}=\text{Max}\left ( \text{JSD}\left ( Y_{i}, Y_{i^{\prime}}\right ) \right )
\end{equation}
where $\text{JSD}\left(Y_{i j}, Y_{i^{\prime} j}\right)$ can be specifically defined as
\begin{equation}
	\begin{aligned}
	&\text{JSD}(Y_{i j} \| Y_{i^{\prime} j})\\
	=&\frac{1}{2} \text{KL}(Y_{i j} \| \frac{Y_{i j}+Y_{i^{\prime} j}}{2})+\frac{1}{2} \text{KL}(Y_{i^{\prime} j} \| \frac{Y_{i j}+Y_{i^{\prime} j}}{2})
	\end{aligned}
\end{equation}

KL divergence is not symmetrical, resulting in two different values for the same two models. So we choose its variant Jensen-Shannon Divergence to measure the diversity between models. In this experiment, we choose two extreme cases in MC-DropReLU method with $q=0.8$ and $q=0.95$.

\begin{table}\footnotesize
	\centering
	\caption{Using Mean Jensen-Shannon Divergence (Mean-JSD) and Max Jensen-Shannon Divergence (Max-JSD) to characterize the diversity of models under the five methods. The \textcolor{red}{red} numbers represent each metric's optimal value, and the \textcolor{blue}{blue} numbers represent each metric's suboptimal value.}
	\begin{tabular}{lcc}
		\toprule
		& \multicolumn{1}{c}{Mean-JSD} & \multicolumn{1}{c}{Max-JSD} \\
		\midrule
		Deep Ensemble & \textcolor{red}{0.020} & \textcolor{red}{0.021} \\
		MC-DropReLU(q=0.8) & \textcolor{blue}{0.017} & \textcolor{blue}{0.017} \\
		SVI & 0.009 & 0.009 \\
		MC-DropReLU(q=0.95) & 0.008 & 0.008 \\
		MC-RReLU & 0.002 & 0.002 \\
		MC-Dropout(p=0.5) & 0.003 & 0.003 \\
		MC-Dropout(p=0.2) & 0.001 & 0.001 \\
		\bottomrule
	\end{tabular}%
	\label{MJSD}%
\end{table}%

As shown in Table \ref{MJSD}, the diversity of Deep Ensemble is the best with Mean-JSD 0.020 and Max-JSD 0.021 respectively, followed by MC-DropReLU(q=0.8) with Mean-JSD 0.017 and Max-JSD 0.017 respectively. This indicates that the prediction results of the models obtained by our proposed sampling method are better than MC-Dropout and slightly worse than Deep Ensemble in terms of the distance metric.

\begin{figure*}
	\centering
	\includegraphics[width=\textwidth]{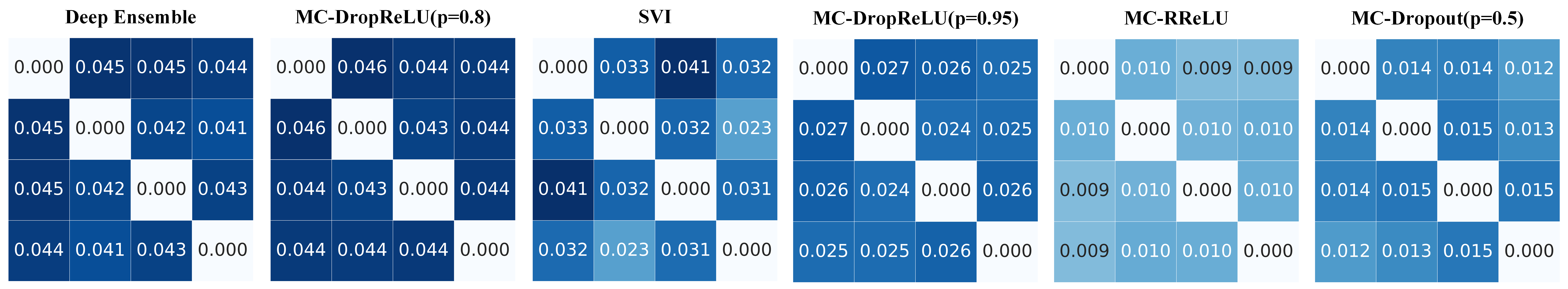}
	\caption{Using The fraction of labels on which the predictions from different checkpoints disagree to characterize the diversity of ResNet-18 under the five methods.}
	\label{DIS}
\end{figure*}

\subsubsection{Disagreement of predictions} \label{4.4.2}

Our goal is to observe the inconsistent results obtained by sampling different models for the same inputs. We consider the disagreement in function space, defined as the fraction of points the checkpoints disagree on, that is:
\begin{equation}
	\text{DIS} = \frac{1}{N} \sum_{n=1}^{N}\left[f\left(x_{n} ; \theta_{1}\right) \neq f\left(x_{n} ; \theta_{2}\right)\right]
\end{equation}
where $f\left(x ; \theta\right)$ denotes the class label predicted by the network for input $x$. In ensemble-based method, each $f$ represents an ensemble element with different initialization. And in sampling-based model, each $f$ represents a network obtained by sampling. In this experiment, we also choose two extreme cases in MC-DropReLU method with $q=0.8$ and $q=0.95$. Figure \ref{DIS} shows the results.

As shown in Table \ref{MDIS}, the diversity of MC-DropReLU (q=0.8) is the best with Mean-DIS 0.044 and Max-DIS 0.046 followed by Deep Ensemble with Mean-DIS 0.043 and Max-DIS 0.045, respectively. This indicates that the prediction results of the models obtained by our proposed sampling method are significantly better than MC-Dropout and slightly better than Deep Ensemble in terms of the disagreement metric.

Combining the above two diversity measurement methods in Section \ref{4.4.1} and Section \ref{4.4.2}, our method is competitive with Deep Ensemble in terms of diversity.

\begin{table}\footnotesize
	\centering
	\caption{Using Mean Disagreement of predictions (Mean-DIS) and Max Disagreement of predictions (Max-DIS) to characterize the diversity of models under the five methods. The \textcolor{red}{red} numbers represent each metric's optimal value, and the \textcolor{blue}{blue} numbers represent each metric's suboptimal value.}
	\begin{tabular}{lcc}
		\toprule
		& \multicolumn{1}{c}{Mean-DIS} & \multicolumn{1}{c}{Max-DIS} \\
		\midrule
		Deep Ensemble & \textcolor{blue}{0.043} & \textcolor{blue}{0.045} \\
		MC-DropReLU(q=0.8) & \textcolor{red}{0.044} & \textcolor{red}{0.046} \\
		SVI & 0.032 & 0.041 \\
		MC-DropReLU(q=0.95) & 0.026 & 0.027 \\
		MC-RReLU & 0.010 & 0.010 \\
		MC-Dropout(p=0.5) & 0.014 & 0.015 \\
		MC-Dropout(p=0.2) & 0.009 & 0.011 \\
		\bottomrule
	\end{tabular}%
	\label{MDIS}%
\end{table}%

\subsection{Position and configuration analysis of MC-DropReLU} \label{4.5}

In this part, we focus on Question 4. When using MC-Dropout in practical applications, where to insert the dropout layers, how many to use, and the choice of dropout rate are often empirically made, leading to possibly suboptimal performance \cite{verdoja2020notes}. We will also face these troubles when using RBUE in this paper. Therefore, in this section, we give a quantitative analysis about where to use the DropReLU layers and the choice of DropReLU rate for reference. By comparing previous experiments, we found that the performance of MC-DropReLU is better than that of MC-RReLU, so the analysis here mainly focuses on MC-DropReLU.

\begin{table}\footnotesize
	\centering
	\caption{Position analysis of MC-DropReLU(0.8) on TinyImageNet with DenseNet on three metrics.}
	\begin{tabular}{lccc}
		\toprule
		& All Layers & Last Layer & First Layer \\
		\midrule
		Accuracy$\uparrow$   & \multicolumn{1}{c}{0.63} & \multicolumn{1}{c}{0.63} & \multicolumn{1}{c}{0.63} \\
		ECE$\downarrow$   & \multicolumn{1}{c}{0.04} & \multicolumn{1}{c}{0.06} & \multicolumn{1}{c}{0.06} \\
		Training Time  & 13.2h & 9.3h  & 9.3h \\
		\bottomrule
	\end{tabular}%
	\label{position}%
\end{table}%

To analyse the influence of the position of DropReLU layers in the neural network, we conduct experiments on TinyImageNet with DenseNet. We divide the placement of DropReLU layer into three cases: \textit{All Layers}, \textit{Last Layer}, and \textit{First Layer}. 'All Layers' means we place DropReLU layers before all the convolutional and fully connected layers. 'Last Layer' means we only place DropReLU layer before the fully connected layer. 'First Layer' means we only place DropReLU layer before the first convolutional layer. As shown in Table \ref{position}, the more DropReLU layers, the greater the diversity of the final results, and the better the model calibration metric ECE. However, the more DropReLU layers mean the increase of sampling times, which will lead to the increase of training time. Moreover, this part of the increased training time will increase with the model and dataset size increase.

\begin{figure}
	\centering
	\includegraphics[scale=1]{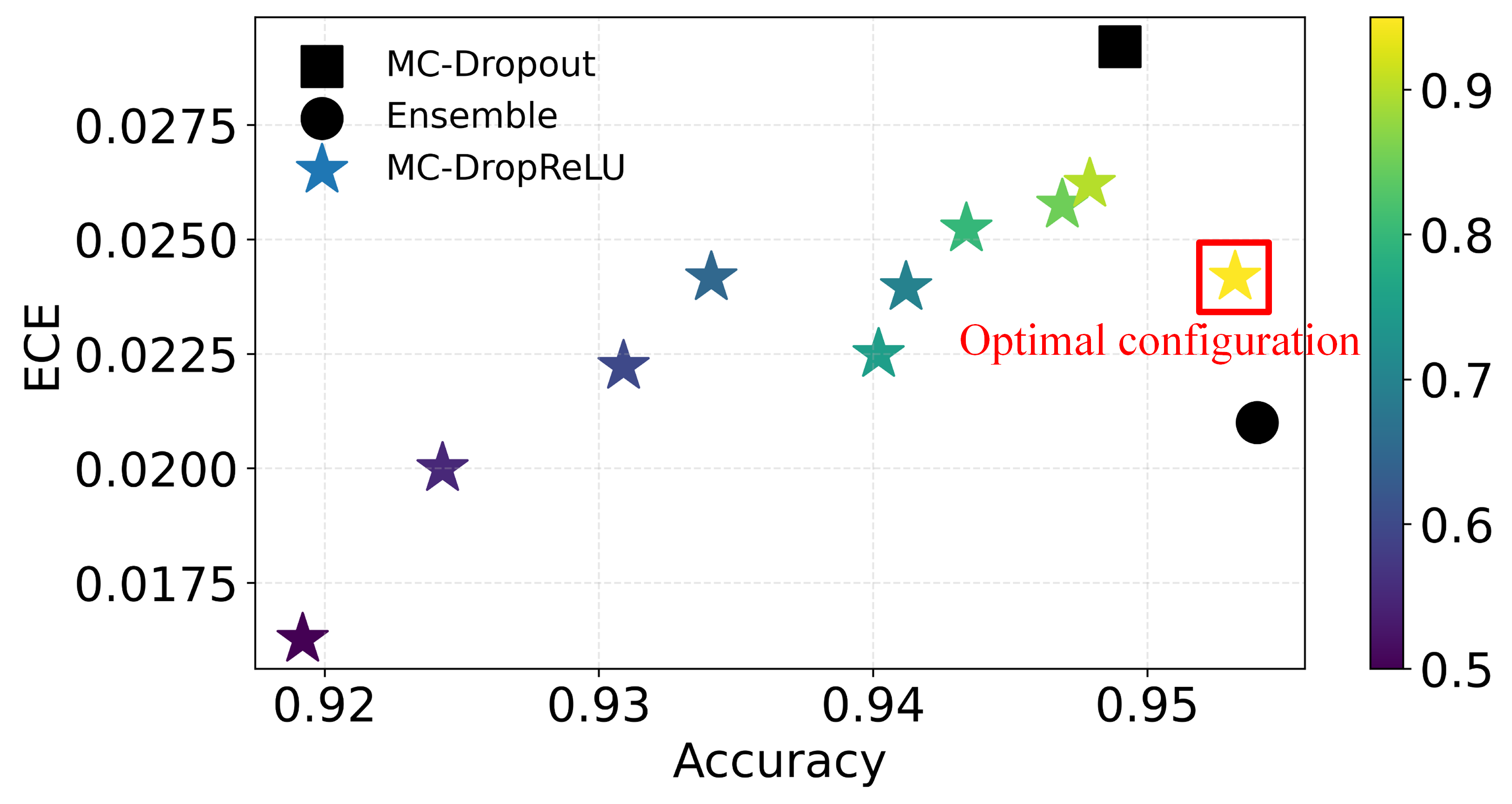}
	\caption{Spanning the space of behaviors. Models in the bottom right cornor are better. The color represents the DropReLU rate $q$.}
	\label{resnet-cifar10-space}
\end{figure}

To analyse the influence of the DropReLU rate, we conduct experiments on ResNet-18 with CIFAR10. Figure \ref{resnet-cifar10-space} depicts the resulting range of behaviors. The 2D coordinates of the markers depict their accuracy and ECE, and their colors correspond to the hyperparameter $q$. For comparison purposes, we also display MC-Dropout and Deep Ensemble results in a similar manner, simply replacing the star with a square and a circle, respectively. As can be seen, the optimal MC-DropReLU configuration depicted by the yellow star can provide better performance than MC-Dropout and performance close to Deep Ensemble. Although the ECE of the configuration depicted by the yellow star is not the smallest, it is the best result after a trade-off between ECE and Accuracy.

\section{Conclusions and future work}

In this work, we introduce RBUE, a novel method to estimate uncertainty in deep neural networks. Instead of using a fixed number of independently trained models as in Deep Ensemble or randomly dropping some neurons at each training step as in MC-Dropout, we propose two strategies MC-DropReLU and MC-RReLU which add randomness to ReLU to get diverse predictions. The main difference between them is the sampling distribution of the slope of the negative semi-axis of ReLU. Furthermore, through the variance analysis of the outputs, we get the selection basis of the hyperparameter in the proposed method and verify this in experiments. Moreover, by changing the hyperparameter $q$, we can span a range of behaviors between those of MC-Dropout and Deep Ensemble. This allows us to identify model configurations that provide a useful trade-off between the high-quality uncertainty estimates of Deep Ensemble at a high computational cost and the lower performance of MC-Dropout at a lower computational cost. Our experiments demonstrate that we can achieve the performance on par with that of Deep Ensemble at a fraction of the cost. In the future, we will apply our method to tasks that require a scalable uncertainty estimation method, in particular active learning and out-of-distribution detection.

\section*{Acknowledgement}

This work was supported by the Natural Science Foundation of China (No. 11725211, 52005505, 62001502) and the Postgraduate Scientific Research Innovation Project of Hunan Province (CX20200006).

\section*{Conflict of interest statement}
The authors declare that they have no conflict of interest.

\bibliographystyle{unsrt}
\bibliography{mybibfile}   

\end{document}